\documentclass[letterpaper]{article} 
\usepackage{aaai2026}  
\usepackage{times}  
\usepackage{helvet}  
\usepackage{courier}  
\usepackage[hyphens]{url}  
\usepackage{graphicx} 
\usepackage{natbib}  
\usepackage{caption} 
\usepackage{algorithm}
\usepackage{algorithmic}
\usepackage{amsmath, amssymb}
\usepackage{subcaption}
\usepackage[varbb]{newtxmath}
\usepackage{array, makecell}   
\setcellgapes{5pt}             

\usepackage{newfloat}
\usepackage{listings}
\DeclareCaptionStyle{ruled}{labelfont=normalfont,labelsep=colon,strut=off} 
\floatstyle{ruled}
\newfloat{listing}{tb}{lst}{}
\floatname{listing}{Listing}
\title{Does Less Hallucination Mean Less Creativity?\\An Empirical Investigation in LLMs}
\author{
    Banerjee Mohor\textsuperscript{\rm 1}\equalcontrib,
    Nadya Yuki Wangsajaya\textsuperscript{\rm 1}\equalcontrib,
    Syed Ali Redha Alsagoff\textsuperscript{\rm 1}\equalcontrib,
    Tan Min Sen\textsuperscript{\rm 2},
    Zachary Choy Kit Chun\textsuperscript{\rm 2},
    Alvin Chan Guo Wei\textsuperscript{\rm 1}
}
\affiliations{
    \textsuperscript{\rm 1}College of Computing and Data Science, Nanyang Technological University\\

    \textsuperscript{\rm 2}Raffles Institution\\
    \{mohor001, nady0006, syedali001\}@e.ntu.edu.sg\\
}

\begin{document}

\maketitle

\begin{abstract}
Large Language Models (LLMs) exhibit remarkable capabilities in natural language understanding and reasoning, but suffer from hallucination: the generation of factually incorrect content. While numerous methods have been developed to reduce hallucinations, their impact on creative generations remains unexplored. This gap is particularly critical for AI-assisted scientific discovery, which requires both factual accuracy and creative hypothesis generation. We investigate how three hallucination-reduction techniques: Chain of Verification (CoVe), Decoding by Contrasting Layers (DoLa), and Retrieval-Augmented Generation (RAG), affect creativity in LLMs. Evaluating multiple model families (LLaMA, Qwen, Mistral) at varying scales (1B - 70B parameters) on two creativity benchmarks (NeoCoder and CS4), we find that these methods have opposing effects on divergent creativity. CoVe enhances divergent thinking, DoLa suppresses it, and RAG shows minimal impact. Our findings provide guidance for selecting appropriate hallucination-reduction methods in scientific applications, where the balance between factual accuracy and creative exploration is crucial.
\end{abstract}


\section{1. Introduction}

The development of Large Language Models (LLMs) is a landmark achievement in the field of Natural Language Processing (NLP). They exhibit unprecedented abilities in natural language understanding \cite{hendrycksMeasuringMassiveMultitask2021} and reasoning \cite{ yufeiNaturalLanguageReasoning2024,kojimaLargeLanguageModels2023, zhao2025surveylargelanguagemodels}. Unfortunately, modern LLMs often suffer from hallucination, the tendency to generate factually incorrect content \cite{Huang_2025, rawteTroublingEmergenceHallucination2023}. As such, there has been significant effort in the field to understand \cite{yaoLLMLiesHallucinations2024, kalaiWhyLanguageModels2025} and combat \cite{gumaan2025theoreticalfoundationsmitigationhallucination} hallucination, especially in high-stakes domains such as AI-assisted scientific discovery where factual reliability is essential \cite{zhang2025advancingscientificmethodlarge}.

However, little is known about how interventions that suppress hallucination affect a model’s creative potential — a crucial ingredient for hypothesis generation and scientific ideation. This question matters because creativity often involves making unconventional connections. For example, scientists sometimes generate useful hypotheses by linking concepts that at first seem unrelated; a process that, in models, can resemble the kind of associative leaps that might otherwise be labeled as hallucination. We adopt the definition of creativity from human psychology \cite{guilfordCreativity1950}, where creativity is divided into \textit{convergent thinking}: solving the problem correctly and within means, and\textit{ divergent thinking}: generation of different ideas. In this paper, we aim to investigate the relationship between hallucination-reduction methods and creativity, shedding light on how factual control interacts with creative reasoning.

\begin{figure*}[h]
    \centering
    \includegraphics[width=0.7\textwidth]{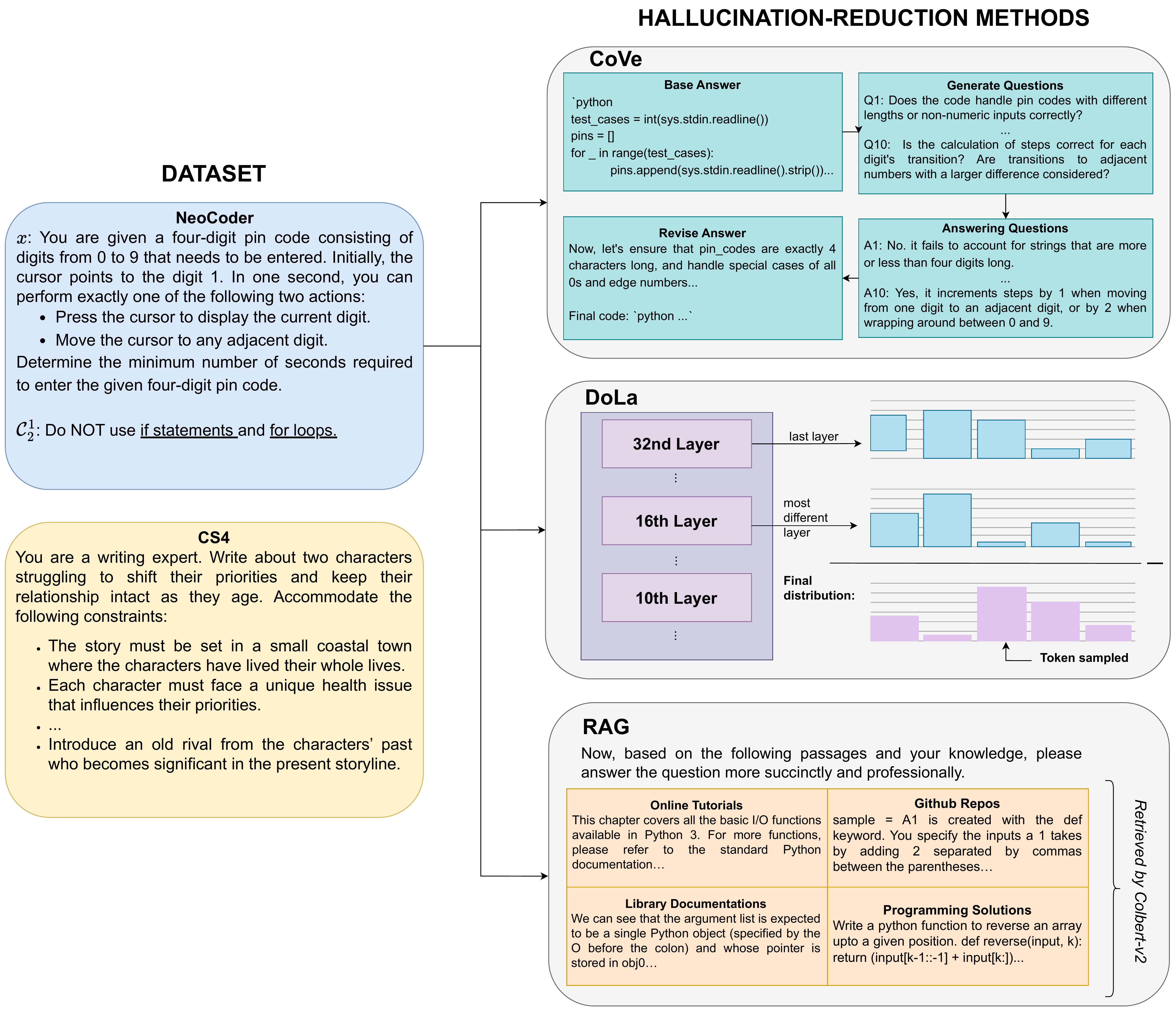}
    \caption{\textbf{Illustration of our experiment framework.} We compare LLM creative performance across two benchmarks (NeoCoder and CS4) with and without three hallucination-reduction methods (CoVe, DoLa, and RAG).}
    \label{fig:pipeline}
\end{figure*}

Our experimental setup is illustrated in Figure \ref{fig:pipeline}. We evaluate creativity using two benchmarks from distinct domains: (1) NeoCoder \cite{lu2025benchmarkinglanguagemodelcreativity}, and (2) CS4 \cite{atmakuru2024cs4measuringcreativitylarge}. NeoCoder evaluates creativity in solving increasingly constrained programming problems: a rule-based setting similar to scientific experimentation under fixed laws. Meanwhile, CS4 tests open-ended story generation, reflecting the imaginative thinking needed for hypothesis generation in science.

We re-implemented three hallucination-reduction techniques: Chain of Verification (CoVe) \cite{dhuliawala2023chainofverificationreduceshallucinationlarge}, Decoding by Contrasting Layers (DoLa) \cite{chuang2024doladecodingcontrastinglayers}, and Retrieval-Augmented Generation (RAG) \cite{lewis2021retrievalaugmentedgenerationknowledgeintensivenlp}. For each, we measure both convergent and divergent creativity before and after applying the method.

Before conducting our experiments, we hypothesized that hallucination-reduction techniques would generally suppress a model’s creative abilities, given that creative thinking often relies on making unconventional associations that may be mistaken for hallucination. However, our empirical results reveal a surprising pattern. Different hallucination-reduction methods affect divergent creativity in opposite ways. CoVe enhances the model’s ability to generate diverse and original ideas, whereas DoLa suppresses it. Meanwhile, convergent creativity is largely unaffected. We further examine whether this trend generalizes across different model families: LLaMA \cite{grattafiori2024llama3herdmodels}, Qwen \cite{hui2024qwen25codertechnicalreport}, and Mistral \cite{jiang2023mistral7b}, and across model scales: 1B, 8B, and 70B parameters. The consistent trend suggests that the creativity-hallucination relationship is an inherent characteristic of LLMs, not an artifact of model size or architecture.

These findings hold notable implications for AI4Science. Scientific discovery depends on maintaining a careful balance between factual accuracy and creative hypothesis generation. Excessive hallucination control may produce models that are precise but limited in imagination, whereas too much generative freedom can lead to factual drift. Our investigation of the creativity-hallucination relationship guides scientists in selecting appropriate hallucination-reduction methods for LLM-driven hypothesis generation.
\bigskip

\noindent In summary, our contributions are as follows.
\begin{enumerate}
    \item {We systematically show that hallucination-reduction methods differentially affect divergent creativity while preserving convergent thinking}
    \item {We show this relationship generalizes across model families (LLaMA, Qwen, Mistral) and scales (1B–70B parameters)}
\end{enumerate}



\section{2. Related Works and Background}
\subsection{Hallucination Reduction Methods}
\paragraph{Chain of Verification} \cite{dhuliawala2023chainofverificationreduceshallucinationlarge} introduces Chain of Verification (CoVe), a structured approach that enhances factual consistency through multi-stage reasoning. The process comprises four stages: (1) drafting an initial answer, (2) generating verification questions based on the draft, (3) answering these questions and synthesizing recommendations, and (4) producing a refined final response. This iterative verification chain enables models to critically evaluate and correct their own outputs before finalizing an answer.

\paragraph{Decoding by Contrasting Layers} \cite{chuang2024doladecodingcontrastinglayers} proposes Decoding by Contrasting Layers (DoLa), a simple decoding method to make large language models more factual. Instead of using only the final layer’s output, DoLa \textit{contrasts} the predictions from a higher layer with those from an earlier one. The earlier “premature” layer is chosen dynamically at each step by finding which layer’s output differs most from the final layer using Jensen-Shannon divergence \cite{61115}. The model then subtracts the earlier layer’s logits from the later layer’s, emphasizing tokens that are learned throughout the layers, while reducing less-reliable tokens from the lower layers.

\paragraph{Retrieval-Augmented Generation} Another widely adopted approach is Retrieval-Augmented Generation (RAG) \cite{lewis2021retrievalaugmentedgenerationknowledgeintensivenlp}, which enhances factual accuracy by retrieving relevant external information from a knowledge source before generating the final response. Integrating retrieved evidence into the model’s context, RAG enables the system to ground its outputs in verifiable data rather than relying solely on parametric memory.

\subsection{Creativity Evaluation Datasets}

\paragraph{NeoCoder}
This dataset \cite{lu2025benchmarkinglanguagemodelcreativity} was introduced to evaluate the creativity of LLMs in a structured, constraint-driven programming setting. It comprises 199 CodeForces problems, each paired with around 30 human-written correct solutions. Every problem $x_i$ is associated with a progressively growing set of constraints $\mathcal{C}_t^i = \{\tau_1^i, \tau_2^i, \ldots, \tau_t^i\}$, where $t$ denotes the constraint state ($t = |\mathcal{C}_t^i|$), and the maximum number of constraints is $T = 5$. Each instance at state $t$ is represented as
\[
\mathcal{D}_t = \{(x_i, \mathcal{C}_t^i)\}_{i=1}^n,
\]
and the corresponding model predictions are obtained as
\[
\mathcal{Y}_t = \{y_i^t \sim \text{LLM}(x_i \oplus \mathcal{C}_t^i), \quad \forall (x_i, \mathcal{C}_t^i) \in \mathcal{D}_t\}.
\]

NeoCoder quantifies creativity along two axes — \textit{convergent} and \textit{divergent} creativity. Using LLM-as-a-Judge, the set of atomic programming techniques $\mathcal{T}_t^i$ employed in each solution $y_i^t$ is extracted. A correctness indicator $\mathbb{1}^{\text{Correct}(y_i^t)}$ equals 1 if all test cases pass. Convergent creativity measures the proportion of correct solutions that simultaneously satisfy all constraints:
\begin{equation}
\begin{aligned}
\textsc{NeoCoder-Convergent}&(\text{LLM}, t) = \\
&\hspace{-5em}\frac{1}{|\mathcal{Y}_t|}
\sum_{y_i^t \in \mathcal{Y}_t}
\mathbb{1}^{\mathcal{T}_t^i \cap \mathcal{C}_t^i = \varnothing}
\, \mathbb{1}^{\text{Correct}(y_i^t)}.
\end{aligned}
\end{equation}

Let $\widehat{\mathcal{T}}^i$ denote all atomic techniques observed in human-written solutions for $x_i$. Divergent creativity captures novelty beyond human patterns:
\begin{equation}
\begin{aligned}
&\textsc{NeoCoder-Divergent}(\text{LLM}, t) = \\
&\hspace{3em}\frac{1}{|\mathcal{Y}_t|}
\sum_{y_i^t \in \mathcal{Y}_t}
\frac{|\mathcal{T}_t^i \setminus \widehat{\mathcal{T}}^i|}{|\mathcal{T}_t^i|}.
\end{aligned}
\end{equation}


Through this formulation, NeoCoder provides a fine-grained measurement of creativity, jointly assessing an LLM’s ability to generate functionally correct with diverse techniques under progressively complex constraints.

\paragraph{CS4}
The CS4 benchmark \citep{atmakuru2024cs4measuringcreativitylarge} provides a controlled framework for evaluating LLMs on creative story generation under progressively more complex constraints. It employs a constraint-generation strategy that produces stylistic and open-ended constraints from user instructions. Each of the 50 instructions is expanded to 39 cumulative constraints, segmented into sets of 7, 15, 23, 31, and 39. This results in 250 unique prompts $(50 \times 5)$.

For every instruction $x_i$ and active constraint set $\mathcal{C}_t^i = \{\tau_1^i, \tau_2^i, \ldots, \tau_t^i\}$, the model generates an output story 
$y_i^t \sim \text{LLM}(x_i \oplus \mathcal{C}_t^i)$. 
The dataset follows a two-stage generation pipeline: GPT-4 first produces a ``base'' story given only $x_i$, and the target LLM revises it to satisfy $\mathcal{C}_t^i$, ensuring models cannot simply restate or memorize constraints.

Each story is evaluated along four quantitative dimensions:

\begin{enumerate}
    \item \textbf{Constraint Satisfaction:} Computed as the proportion of fulfilled constraints, automatically judged by LLM-as-a-Judge:
    \begin{equation}
        \text{Constraint Satisfaction} = 
        \frac{\text{\# of satisfied constraints}}{\text{Total constraints}}.
    \end{equation}

    \item \textbf{Coherence:} Measured via pairwise LLM-as-a-Judge comparisons against a baseline story (at 23 constraints), rated from 1-5 and normalized to [0,1]:
    \begin{equation}
        \text{Coherence}_{\text{norm}} = 
        \frac{\text{Mean Coherence Score}}{5}.
    \end{equation}

    \item \textbf{Diversity:} Quantified by \textsc{Dist-n} diversity, defined as the product of unique $n$-gram ratios:
    \begin{equation}
        \textsc{Dist-n} = 
        \prod_{n=2}^{4} 
        \frac{|\text{unique $n$-grams}|}{|\text{total $n$-grams}|}.
    \end{equation}

    \item \textbf{Creativity:} Evaluated through a composite metric:
        \begin{equation}
            \textsc{QUC}_n = 
            (\text{Coherence}_{\text{norm}}) 
            \times 
            (\text{Constraint Satisfaction}) \label{eq:QUC}
        \end{equation}
        where $\textsc{QUC}_n$ (Quality Under $n$ Constraints) captures story quality at constraint level $n$.
\end{enumerate}

Similar to NeoCoder, using this framework, CS4 also assesses LLMs ability to balance correctness and diversity under increasing task complexity.

\section{3. Effects of Hallucination-Reduction Methods on Creativity}
\begin{table*}[!ht]
\centering
\small
\makegapedcells  
\begin{tabular}{|l|l|l|}
\hline
\multicolumn{1}{|c|}{\textbf{Metric}} & \multicolumn{1}{c|}{\textbf{NeoCoder}} & \multicolumn{1}{c|}{\textbf{CS4}} \\ 
\hline
Convergent Creativity & $\text{\textsc{NeoCoder}-Convergent}$ & \textsc{QUC} \\
Divergent Creativity & $\text{\textsc{NeoCoder}-Divergent}$ & Diversity (\textsc{Dist-N}) \\

  
\hline
\end{tabular}
\caption{List of metrics for creativity evaluation.}
\label{table:metrics}
\end{table*}

\begin{figure*}[!h]
    \centering
    \includegraphics[width=\textwidth]{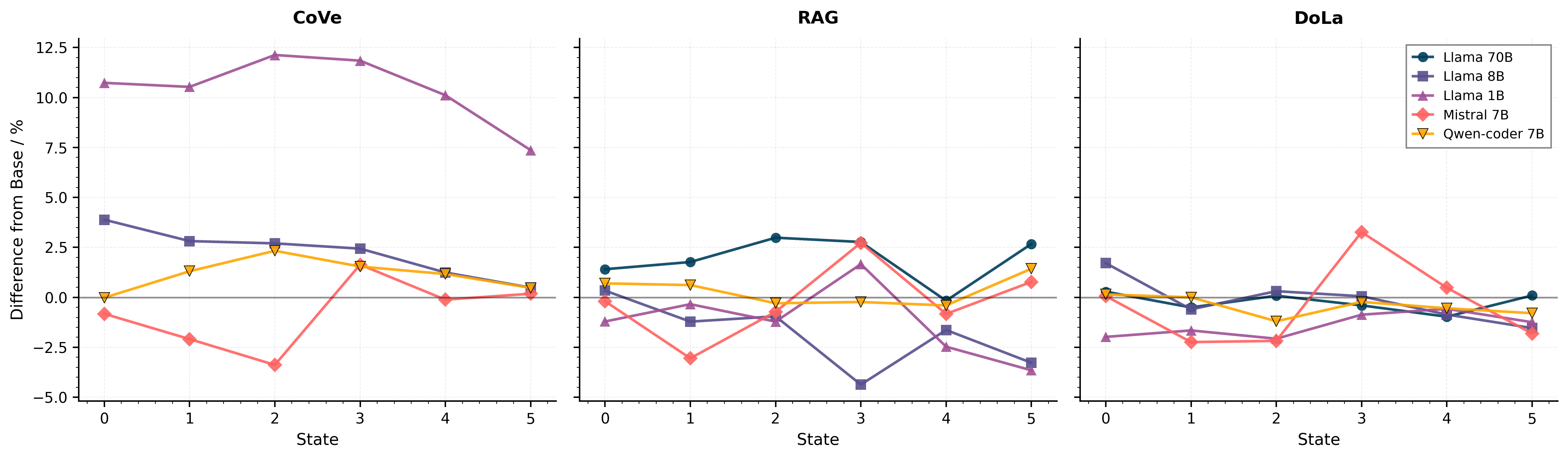}

    \caption{\textbf{Impact of decoding methods on divergent creativity (NeoCoder).} The plots show the percentage improvement over baseline performance for various language models across six constraints. The horizontal line at y=0 represents the baseline (generations without hallucination-reduction methods). Positive values indicate improvement over baseline, while negative values indicate degradation.}
    \label{fig:hal-methods-div-neocoder}
\end{figure*}

\begin{figure*}[!h]
    \centering
    \includegraphics[width=0.6667\textwidth]{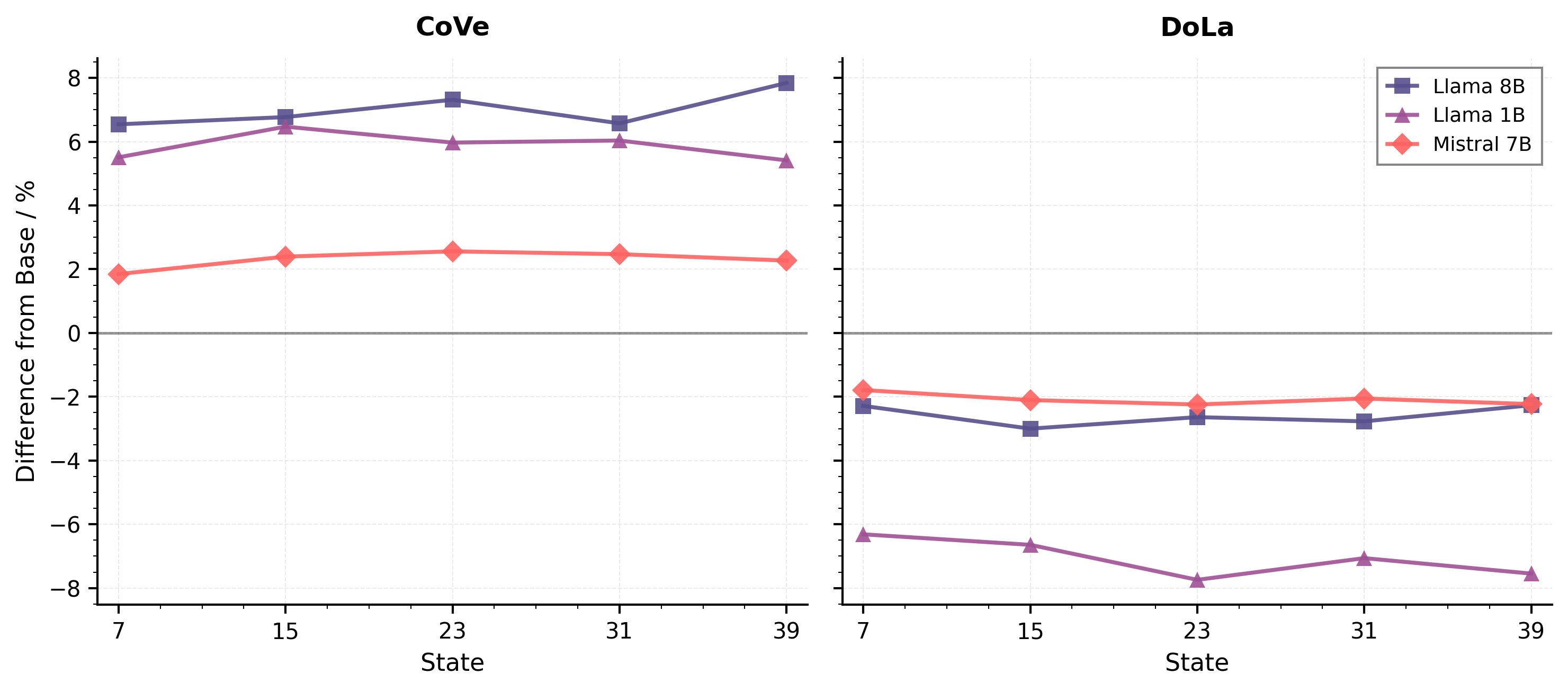}
    \caption{\textbf{Impact of decoding methods on divergent creativity (CS4).} The plots show the percentage improvement over baseline performance for various language models across 39 constraints. The horizontal line at y=0 represents the baseline (generations without hallucination-reduction methods). Positive values indicate improvement over baseline, while negative values indicate degradation.}
    \label{fig:hal-methods-div-cs4}
\end{figure*}

\subsection{Experimental Setup}
In this section, we provide details on the structure of our experiment (Figure \ref{fig:pipeline}).
We evaluated models on the two datasets — NeoCoder \cite{lu2025benchmarkinglanguagemodelcreativity} and CS4 \cite{atmakuru2024cs4measuringcreativitylarge} — to assess both factual consistency and creative robustness. For the NeoCoder benchmark, we evaluated five models: LLaMA 70B, LLaMA 8B, LLaMA 1B \cite{grattafiori2024llama3herdmodels}, Mistral 7B \cite{jiang2023mistral7b}, and Qwen-Coder 7B \cite{hui2024qwen25codertechnicalreport} to ensure comprehensive coverage across diverse model families and parameter scales. For the CS4 benchmark, experiments were conducted with LLaMA 8B, LLaMA 1B, and Mistral 7B under identical experimental settings for consistency. All generations were performed three times independently using identical configurations, and the mean performance across runs was reported for reliability. Wherever LLM-as-a-Judge evaluation was required, GPT-5-mini \cite{openai2025gpt5} was used for consistency and neutrality.

We re-implemented all three hallucination-reduction methods using the following settings. CoVe was implemented using the AutoGen multi-agent framework \cite{wu2023autogenenablingnextgenllm}, while for DoLa, we chose to only contrast with even-indexed layers, as per the original implementation. Finally, for RAG, we employed ColBERTv2 \cite{santhanam2022colbertv2effectiveefficientretrieval}, following the RAGLAB Framework \cite{zhangRAGLABModularResearchOriented2024}, as the retrieval backbone. It indexed a large corpus of coding tutorials, library documentations, Github repositories, and programming solutions, taken from the CodeRAG-bench \cite{wangCodeRAGBenchCanRetrieval2025} retrieval documents. Documents were vectorized into ColBERTv2 embeddings and retrieved via cosine similarity, with the top 3 ranked segments appended to the prompt for context-grounded generation. RAG was applied only to the NeoCoder benchmark, as open-ended story generation in CS4 lacks a defined retrieval corpus. Further fine-grained details are available in Appendix E.

\subsection{Results}
We used the metrics summarized in Table 1 to evaluate models' performance on the NeoCoder and CS4 benchmarks. 
For convergent and divergent creativity, we followed the definitions provided in the respective benchmark papers (\textsc{NeoCoder-Convergent} / \textsc{Neocoder-Divergent} for NeoCoder and \textsc{QUC} / \textsc{Dist-n} for CS4). 
To quantify the impact of each hallucination-reduction method, we report the percentage changes relative to baseline performance, where the baseline represents generation without any hallucination-reduction methods. Formally, for a given metric $M$, the percentage improvement is calculated as:
\begin{equation}
\text{Difference from Base}(\%) = \frac{M_{\text{method}} - M_{\text{baseline}}}{M_{\text{baseline}}} \times 100
\end{equation}
where $M_{\text{method}}$ is the metric value when using a specific hallucination-reduction method (CoVe, DoLa, or RAG), and $M_{\text{baseline}}$ is the metric value for generation without any hallucination-reduction methods. Positive values indicate a performance improvement, while negative values indicate a performance degradation relative to the baseline.


In the following sections, we focus on divergent creativity, as our observations indicate hallucination-reduction methods mainly impact this dimension, leaving convergent creativity relatively unaffected. Further analysis on convergent creativity is available in Appendix A.

\paragraph{CoVe increases divergent creativity}
As shown in Figure \ref{fig:hal-methods-div-neocoder} and \ref{fig:hal-methods-div-cs4}, CoVe decoding enhances divergent creativity across most evaluated models. On the NeoCoder dataset (Figure \ref{fig:hal-methods-div-neocoder}), LLaMA 1B achieves the highest improvement, peaking at around 12.5\% above baseline, while LLaMA 8B and Qwen-coder 7B show moderate gains of 2–4\%. Mistral 7B, however, represents a deviation from this overall upward trend, with values ranging from –3\% to +2\% relative to the baseline. On the CS4 dataset (Figure \ref{fig:hal-methods-div-cs4}), LLaMA 8B and LLaMA 1B maintain steady improvements of approximately 5–8\%, and Mistral 7B shows a modest but stable increase of around 2\%. Overall, CoVe demonstrates consistent improvements in divergent creativity across models and datasets. 

Our observation supports the hypothesis that questioning improves creativity \cite{wroblewskaApplyingTextMining2025}, by encouraging a broader exploration of the solution space in the model. This aligns with recent findings that exploration of different reasoning paths can help models sidestep 'tunnel vision' \cite{wen2025parathinker} and explore better, more unique solutions to problems.  In human cognition, similar mechanisms are well-documented: questioning strategies reduce fixation and enhance creative output \cite{razRoleAskingMore2025}, while brainstorming techniques stimulate divergent thinking \cite{ritter2017enhancementcreativethinking}. The CoVe verification process may function analogously, prompting the model to reconsider and explore alternative solutions rather than committing prematurely to a single response path.

\paragraph{RAG has no effect on divergent creativity}

Across all evaluated models in Figure \ref{fig:hal-methods-div-neocoder} and \ref{fig:hal-methods-div-cs4}, RAG generations show minimal influence on divergent creativity. LLaMA 70B exhibits small positive shifts up to about +3\%, while LLaMA 8B shows a decline, dropping to around –5\% with no positive deviation. Qwen-coder 7B remains close to baseline, moving between roughly –0.5\% and +1.5\%. LLaMA 1B varies between approximately –3\% and +1.5\%, and Mistral 7B ranges from about –2.5\% to +2.5\%. These model-specific fluctuations, showing both positive and negative shifts, indicate that RAG does not meaningfully influence the models’ creative performance. 

This neutral effect could stem from retrieval quality issues. Studies show that irrelevant retrieved documents introduce noise that misleads LLM generation \cite{shi2023large}, showing that performance could degrade when context lacks direct relevance to the query. Recent work has also shown that redundant knowledge in RAG corpora can hurt performance on questions the LLM can already answer \cite{luo2024zerorag}. In our setup, CodeForces problems are presented through narrative scenarios, while the retrieval corpus contains technical coding tutorials and documentation from CodeRAG-bench. This semantic mismatch between anecdotal problem descriptions and tutorial-style documentation results in retrieved documents that lack the specific algorithmic insights needed for competitive programming tasks. However, this net-neutral effect suggests potential for improvement: if RAG were to retrieve documents containing newer patterns or problem-solving strategies outside the model's training distribution, it could enhance both convergent creativity (by providing relevant factual guidance) and divergent creativity (by exposing the model to unfamiliar approaches). The key lies in ensuring retrieved documents offer genuinely new and relevant information rather than redundant or misaligned content.

\paragraph{DoLa reduces divergent creativity}

Figure \ref{fig:hal-methods-div-neocoder} and \ref{fig:hal-methods-div-cs4} show that DoLa decoding leads to a slight reduction in divergent creativity across both datasets and most models. In the NeoCoder dataset, most models perform slightly below the baseline, with LLaMA 1B ranging from approximately –2.5\% to –1\%, and Mistral 7B remaining mostly below baseline, reaching around –2.5\% with only a single rise to about +3\% at the third state. LLaMA 8B, Qwen-coder 7B, and LLaMA 70B remain nearly constant, with values between –1\% and –0.5\%. In the CS4 dataset, the reduction is more pronounced, as LLaMA 1B drops to around –8\%, while LLaMA 8B and Mistral 7B show smaller decreases of roughly –3\% to –2\%.

The consistency of this divergent creativity-dampening effect across model families and parameter scales suggests that the phenomenon is fundamental to the DoLa approach rather than an artifact of model architecture or size. Overall, the results indicate that DoLa systematically dampens divergent creativity across diverse tasks.

We hypothesize that this phenomenon is caused by DoLa inadvertently contrasting layers responsible for creativity. Since DoLa works by subtracting early-layer predictions from late-layer predictions to enhance factuality, and if early layers encode more exploratory and divergent representations, this contrastive operation may suppress the very features necessary for creative generation. This line of thinking led us to investigate which specific layers correlate with creativity and to experiment with reversing DoLa's effect to improve rather than reduce creativity.


\begin{figure*}[!h]
\centering 
{%
\includegraphics[width=0.7\textwidth]{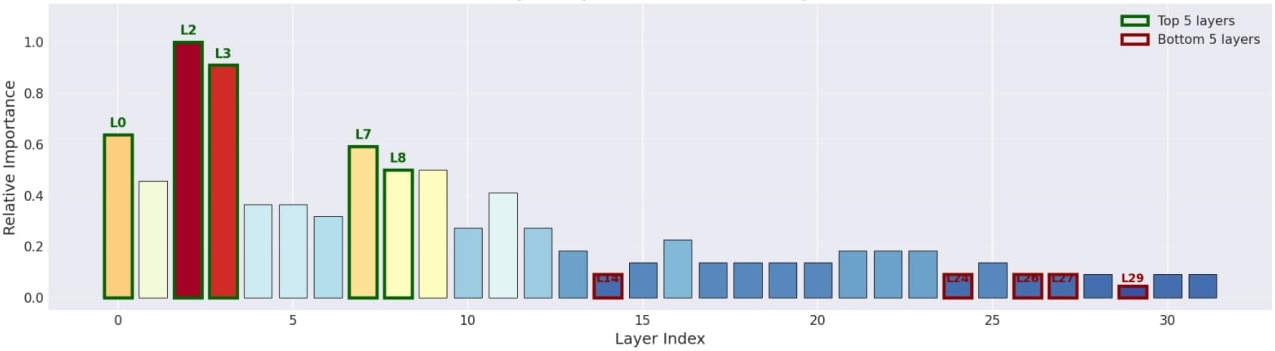}
  \includegraphics[width=0.7\textwidth]{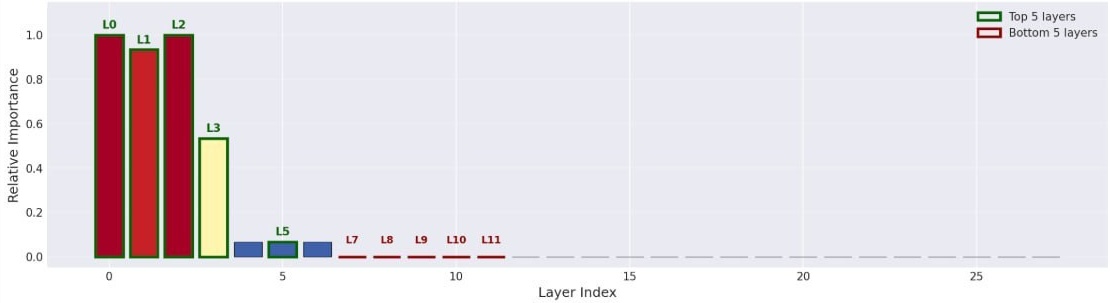}%
}
\caption{How well the linear probes attached to each layer predicts creativity. The y-axis is normalized to the highest value. The top 5 layers, with green borders, are the \textit{creativity-correlated layers}. Meanwhile, the bottom 5 layers, with red borders, are the \textit{anti-correlated layers}. As the \textit{creativity-correlated layers} often cluster at early layers, this shows that \textbf{early layers play a large role in predicting creativity}.}
\label{fig:probes}
\end{figure*}

\begin{figure*}[!h]
    \centering
    \includegraphics[width=0.667\textwidth]{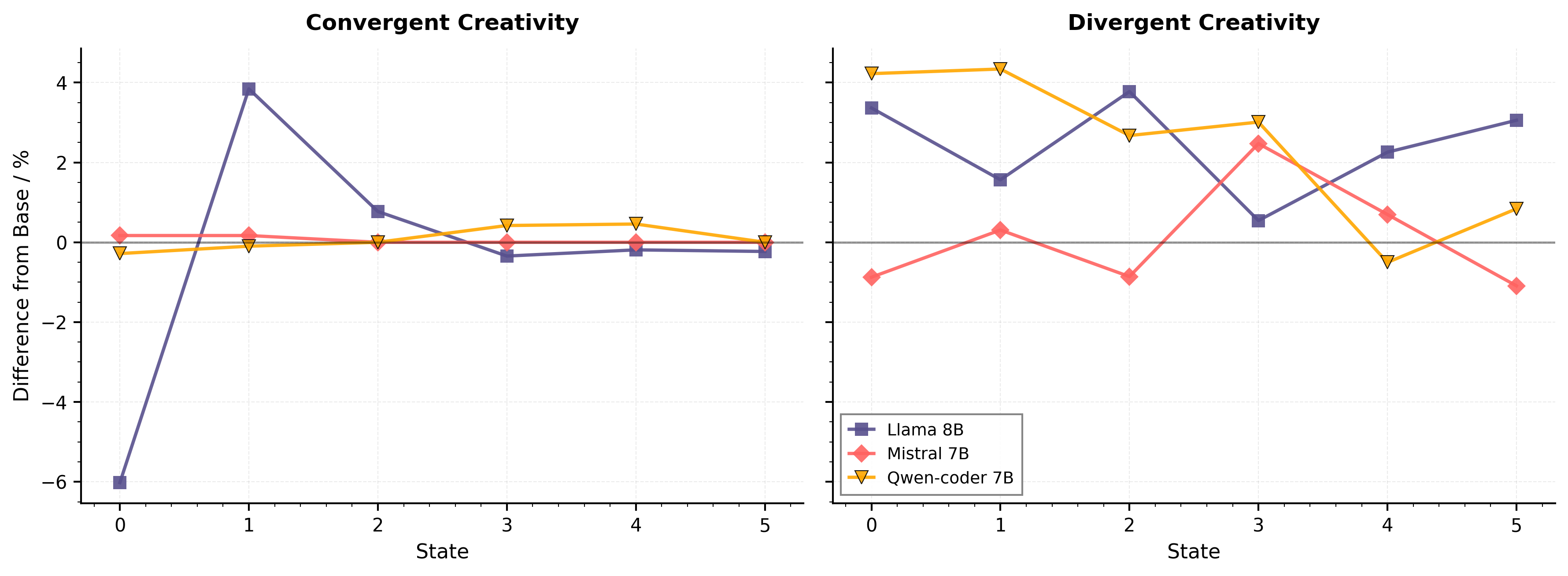}
    \includegraphics[width=0.667\textwidth]{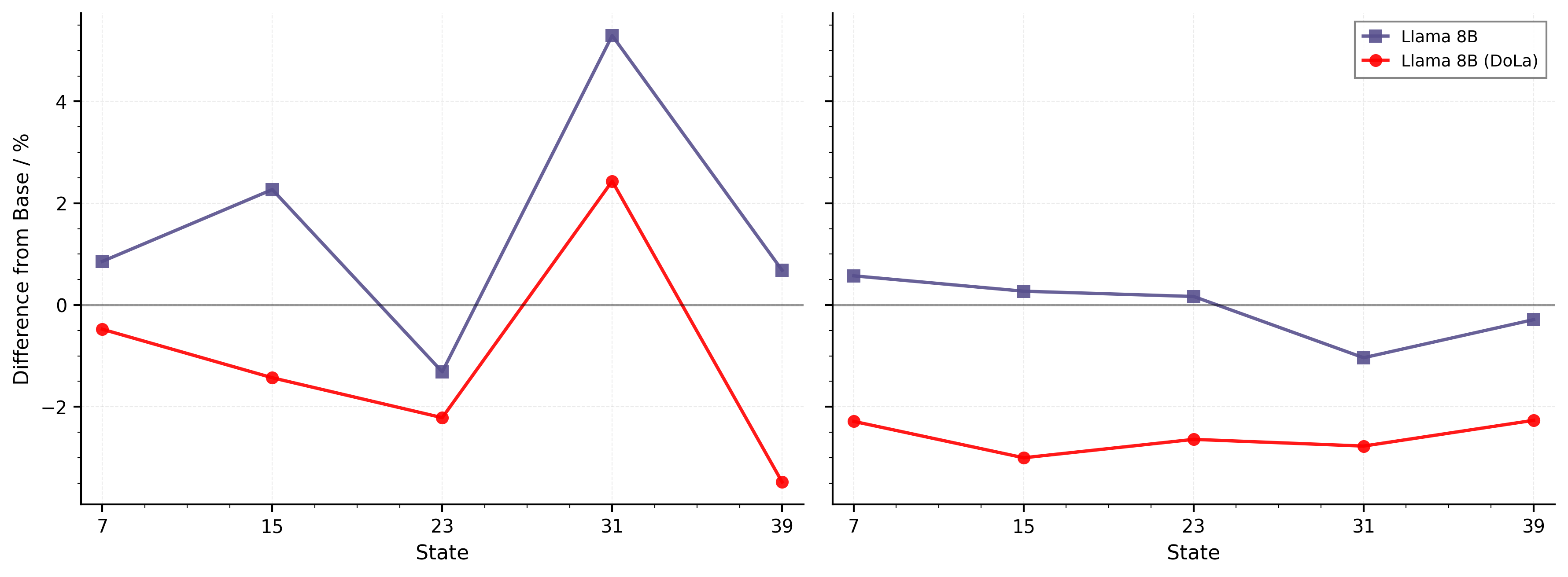}
        \label{fig:tok-sub-div-con-b}
    
    \caption{Enhancing divergent creativity by amplifying creativity-correlated layers and suppressing anti-correlated layers. Top: NeoCoder dataset. Bottom: CS4 dataset. \textbf{This method boosts divergent creativity (right panels) without compromising convergent creativity (left panels)} in both datasets. Note that CS4 results are evaluated on LLaMA 8B only due to computation constraints.}
    \label{fig:tok-sub-div-con}
\end{figure*}

\subsection{Further Studies on DoLa}
We investigated the influence of early layers on the model's divergent creativity. Using probing methods inspired by Inference-Time Intervention (ITI) \cite{liInferenceTimeInterventionEliciting2024}, we used linear probes to identify which attention head are most correlated with creativity. Specifically, we trained linear probes to predict whether the model is going to generate a divergently creative output on the NeoCoder dataset. Since DoLa operates on entire layers rather than individual attention heads, we aggregated the attention head-level correlations within each layer to obtain layer-level correlation scores for direct compatibility. Figure \ref{fig:probes} shows that early layers exhibit stronger correlations with creativity than later layers. This finding supports our hypothesis that DoLa's contrastive decoding mechanism, which specifically contrasts against early layer representations, inadvertently suppresses divergent creativity by removing the very layer activations responsible for creative generation.

Building on this finding, we show promise in enhancing divergent creativity through targeted layer modulation. We define the top 5 layers with the strongest positive correlations as \textit{creativity-correlated layers} and the bottom 5 as \textit{anti-correlated layers} (Figure \ref{fig:probes}). By amplifying creativity-correlated layers while suppressing anti-correlated layers during decoding, we could boost divergent creativity in LLaMA and Qwen-coder (Figure \ref{fig:tok-sub-div-con}) without compromising convergent creativity. The improvement for the NeoCoder dataset in Figure \ref{fig:tok-sub-div-con} is compared with Figure \ref{fig:hal-methods-div-neocoder} (DoLa). This dissociation reveals that divergent and convergent creativity are possibly decoupled, making it possible to enhance one without degrading the other. Implementation details are provided in the Appendix B. We also provide more insight into this divergent creativity-improving method in Appendix C and D.

\section{5. Limitations}
We chose to evaluate creativity using programming (NeoCoder) and story generation (CS4) benchmark, as they both offer established metrics for measuring convergent and divergent creativity. While programming tasks provide rule-based constraints analogous to scientific laws, and story generation reflects open-ended ideation, they are only a proxy of actual scientific hypothesis generation. Future work should develop creativity evaluation frameworks specifically for scientific hypotheses to determine whether our observed creativity-hallucination relationships persist in authentic scientific discovery contexts.

Furthermore, we show that CoVe consistently enhances divergent creativity across models and datasets. However, we did not conduct ablation studies to isolate the mechanism responsible for this improvement. Systematic investigation of the specific mechanism on why questioning increases divergent creativity remains an important direction for future work.

Similarly, we also show RAG has minimal impact on divergent creativity, which we attribute to potential retrieval irrelevance. However, we did not systematically measure retrieval quality or explore alternative retrieval strategies, beside cosine-similarity. We leave further investigation on this result for future work.



\section{6. Conclusion}
In this paper, we investigated how three hallucination-reduction methods: CoVe, DoLa, and RAG, affect creativity in large language models. Testing across multiple model families (LLaMA, Qwen, Mistral) and scales (1B–70B parameters) on two benchmarks (NeoCoder and CS4), we found that these methods have opposing effects on divergent creativity while leaving convergent creativity largely unchanged.

CoVe enhances divergent creativity across most models and settings. DoLa consistently suppresses it. RAG has minimal effect, likely due to poor retrieval quality in our experimental setup. These different effects matter because they represent different trade-offs between factual accuracy and creative generation.

We used linear probes to investigate why DoLa reduces creativity. Early transformer layers showed stronger correlations with creative output than later layers. Our hypothesis is that since DoLa contrasts early and late layers to improve factuality, it inadvertently suppresses creativity-related representations. We provide a promising preliminary results: by reversing this approach; amplifying creativity-correlated layers while suppressing anti-correlated ones, we improved divergent creativity without harming convergent performance.

As LLMs continue to grow smarter, unlocking their potential for scientific discovery becomes increasingly significant. Our investigation into their creativity and hallucination offers a step toward this direction. We hope to see a future where LLMs act not merely as passive tools, but as active collaborators in scientific ideations.



\section{Acknowledgments}
We are grateful for the support of College of Computing and Data Science in Nanyang Technological University, as well as the CN Yang Scholars Programme. 


\bibliography{aaai2026}
\clearpage
\appendix
\begin{figure*}[h]
    \centering
    \includegraphics[width=\textwidth]{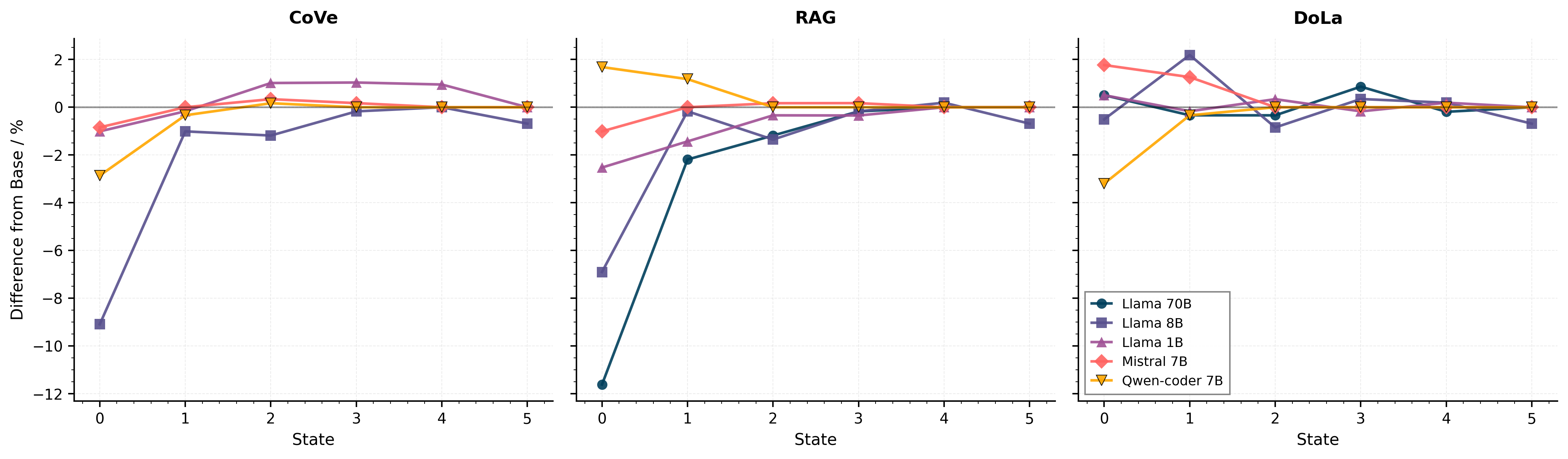}
    \includegraphics[width=0.6667\textwidth]{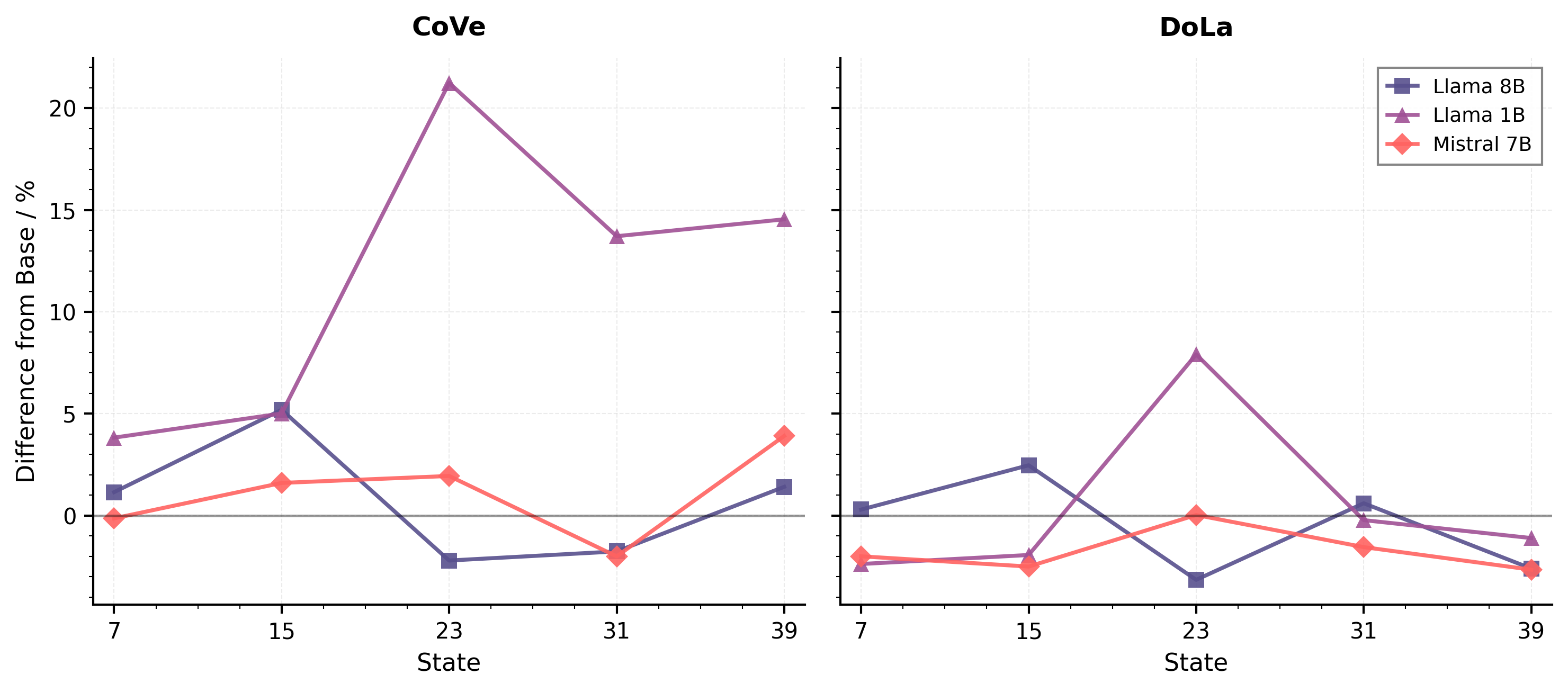}
    \caption{\textbf{Impact of decoding methods on convergent creativity.} The plots show the percentage improvement over baseline performance for various language models across six constraints. Top: NeoCoder dataset. Bottom: CS4 dataset. The horizontal line at y=0 represents the baseline (generation without hallucination-reduction methods). Positive values indicate improvement over baseline , while negative values indicate degradation.}
    \label{fig:hal-methods-cov}
\end{figure*}

\begin{figure*}[h]
    \centering
    \includegraphics[width=0.5\textwidth]{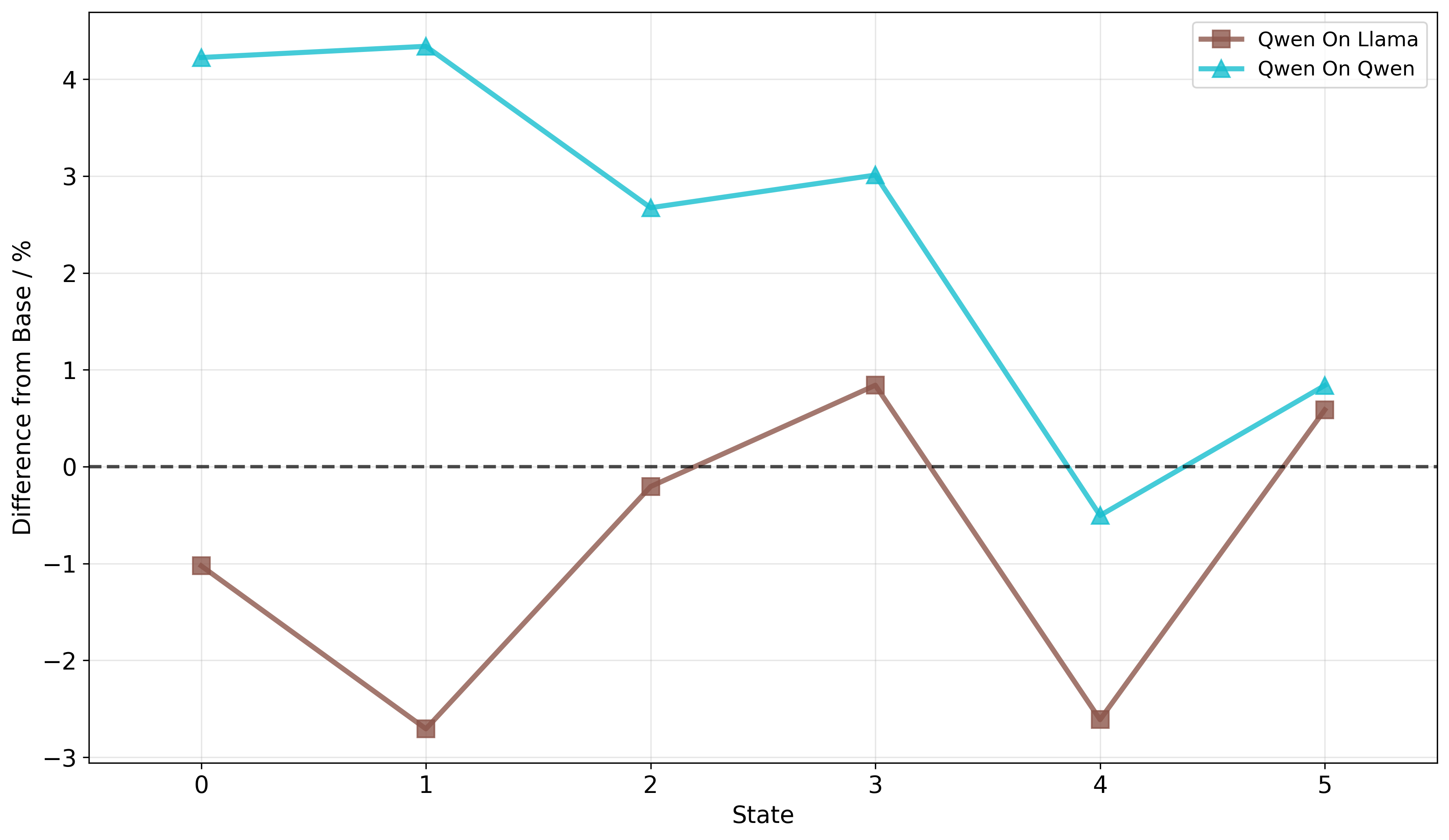}
    \caption{Model-specific nature of creative probes. Divergent creativity improvement across constraint states for Qwen-coder 7B using Creative DoLa with probes trained on different generations. When probes are trained on LLaMA 8B creative generations (blue line), they fail to improve Qwen-coder's divergent creativity. However, when probes are trained on Qwen-coder's own creative generations (brown line), the method successfully enhances divergent creativity. \textbf{This demonstrates that creative probes must be trained on model-specific activations to be effective}.}
    \label{fig:bad-results}
\end{figure*}

\begin{figure*}[!h]
    \begin{align}
    &\textsc{Increasing divergent creativity with probes}(x_{t}|x_{<t}) = \text{softmax}(\mathcal{F}(x_t, N, M, \mathcal{A}, \mathcal{B})), \quad \text{where} \label{eq:creative_dola_equation} \\
    &\mathcal{F}(x_t, N, M, \mathcal{A}, \mathcal{B}) = \begin{cases}
    \overbrace{\log\left( \frac{q_{N}(x_{t})}{q_{M}(x_{t})} \right) }^{\text{Normal DoLa}} + \alpha \left( \sum_{a \in \mathcal{A}}\gamma_{a} \log\left( q_{a}(x_{t}) \right) -  \sum_{b \in \mathcal{B}}\gamma_{b} \log\left(q_{b}(x_{t}) \right) \right), & \text{if } x_{t} \in \mathcal{V}_{\text{head}}(x_{t}|x_{<t}) \\
    - \infty, & \text{otherwise.}
    \end{cases} 
    \end{align}
\end{figure*}
\section{A. Convergent Creativity on Hallucination-Reduction Methods}
As shown in Figure 6, the application of hallucination-reduction methods such as CoVe, RAG, and DoLa does not substantially affect convergent creativity across both datasets and all models. Performance differences relative to the baseline remain minimal, with most values fluctuating around zero. Although minor variations appear at certain states, these deviations do not exhibit a consistent trend. Overall, the results indicate that hallucination-reduction methods preserve convergent creativity, suggesting that such techniques neither enhance nor impair this aspect of model performance.

\section{B. Training Linear Probes}
Following the success of linear probes in identifying properties such as truthfulness \cite{liInferenceTimeInterventionEliciting2024} and even safety-related concepts \cite{teo2025blessingcursedimensionalitysafety}, we similarly hypothesize that creativity-related features may be linearly separable in the representation space.

As for the dataset used to train the linear probes, we curate only convergently creative answers to ensure that the probes learn to distinguish meaningful creative responses, rather than coherent versus incoherent outputs. However, due to limited convergently creative outputs from each individual models, we augmented our dataset by leveraging outputs from hallucination reduction methods. As a form of bootstrapping to include these results, we used partial outputs as a conditioning context to provide a signal for creativity-related activations. Using validation from a small subset of the NeoCoder dataset, we found that using about 40\% of the output as an input signal to the model showed best results. 

\section{C. Details on Divergent Creativity-Improving Method}
We follow similar notation as the original DoLa paper \cite{chuang2024doladecodingcontrastinglayers}.

For a transformer with $N$ layers and input of $t-1$ tokens, we define the output of the $i^{\text{th}}$ layer, where $i \in \{1,2, \dots, N\}$ as $H_i = \{ h_1^{(i)}, h_2^{(i)}, \dots, h_{t-1}^{(i)} \}$. We also have $\phi(\cdot)$ which predicts the probability of next token $x_t$ over the vocabulary $\mathcal{X}$.

Finally, we define \textit{early exit}; instead of applying $\phi(\cdot)$ to the last layer $N$, we apply it to layer $j \in \{1,2, \dots, N-1\}$, to get the output probability $q_j$.
$$
q_j(x_t|x_{<t}) = \text{softmax}(\phi(h_t^{(j)}))_{x_t}
$$
The top 5 layers with strongest correlation to creativity, which we call \textit{creativity-correlated layers}, are placed in set $\mathcal{A}$, while the bottom 5 layers, the \textit{anti-correlated layers}, are placed in set $\mathcal{B}$. The original DoLa implementation also search for layer $M$, which is the layer with highest Jensen-Shannon Divergence compared to layer $N$. The detailed equation for our method is in Equation \ref{eq:creative_dola_equation}. 

Following Contrasting Decoding method \cite{li2023contrastivedecodingopenendedtext}, the subset $\mathcal{V}_{\text{head}}(x|x_t) \in \mathcal{X}$ contains tokens with sufficiently 'high' probabilities at the last layer $N$. The parameter $\beta$ corresponds to the number of tokens considered. In the paper, $\beta$ is set to $0.1$. 
$$
\mathcal{V}_{\text{head}}(x_t|x_{<t}) = \left\{x_t \in \mathcal{X}: q_N(x_t) \geq \beta \max_w q_N(w) \right\}
$$

Furthermore, we set $\alpha = 1$ and all values of $\gamma = 0.5$ due to computation constraints.

\section{D. Creative Probes are Model-Specific}
Creative DoLa interventions only affect the model's own generations, and consequently, our bootstrapping method using partial outputs as conditioning signals is model-specific. The linear probes must be trained on activations from the same base model they will be applied to.

Initially, we trained probes on a dataset containing generations from only LLaMA 3.1 8B and tested them on Qwen-coder and Mistral, which yielded poor results. However, after replacing the training dataset with generations only from those specific models, we observed substantial improvements in performance (Figure \ref{fig:bad-results}).

This model-specific requirement can be explained by distribution shift in activation space. Different language models, even within the same family, learn distinct internal representations and may exhibit different activation patterns for the same inputs. When we extract activations from partial outputs as conditioning signals, we are capturing model-specific patterns of how creativity manifests in that particular model's hidden states. 

Our method conditions the model on partial outputs before extracting activations for probe training. This creates a dependency on both how the model generates these partial outputs, and how it represents the resulting conditioned state internally. Both factors are model-specific, making the extracted activation distribution tied to the source model.

\section{E. Detailed Settings for the Experiments}
All experiments followed standardized decoding settings: `do\_sample=True` (non-greedy decoding to allow natural variation in outputs), `max\_new\_tokens=800` (to enable complete, unconstrained responses), `num\_beam\_groups=1` and `num\_beams=1` (to disable beam search and reflect single-pass reasoning), `temperature=1.0` (set to a high value to encourage more diverse and less deterministic generations), and `top\_p=1.0` (to include the full token distribution). These uniform parameters ensure comparability across all methods and prevent bias.


\end{document}